\documentclass{article}

\usepackage[final]{corl_2020} 
\usepackage{enumitem}
\usepackage{amsmath}

\usepackage{float}
\usepackage{subfigure}
\usepackage{graphicx}
\usepackage{wrapfig}
\usepackage{amsfonts}
\usepackage{stmaryrd}
\usepackage{textcomp}
\usepackage{booktabs}
\usepackage{colortbl}
\usepackage{multirow}
\usepackage{cleveref}
\crefname{figure}{Fig.}{Figs.}
\crefname{table}{Tab.}{Tabs.}
\crefname{equation}{Eq.}{Eqs.}
\crefname{section}{Sec.}{Secs.}

\title{Attentional Separation-and-Aggregation Network \\for Self-supervised Depth-Pose Learning \\in Dynamic Scenes} 

\author{
	Feng Gao$^1$, Jincheng Yu$^1$, Hao Shen$^2$, Yu Wang$^1$, Huazhong Yang$^1$ \\
	{\small \{gao-f19, yjc16\}@mails.tsinghua.edu.cn, shenhao04@meituan.com} \\
	{\small \{yu-wang, yanghz\}@tsinghua.edu.cn}\\
	$^1$ Tsinghua University $^2$ Meituan-Dianping Group \\
}

\begin{document}
\maketitle


\begin{abstract}
	Learning depth and ego-motion from unlabeled videos via self-supervision from epipolar projection can improve the robustness and accuracy of the 3D perception and localization of vision-based robots. However, the rigid projection computed by ego-motion cannot represent all scene points, such as points on moving objects, leading to false guidance in these regions. To address this problem, we propose an \textbf{A}ttentional \textbf{S}eparation-and-\textbf{A}ggregation \textbf{Net}work (\textbf{ASANet}), which can learn to distinguish and extract the scene's static and dynamic characteristics via the attention mechanism. We further propose a novel MotionNet with an ASANet as the encoder, followed by two separate decoders, to estimate the camera's ego-motion and the scene's dynamic motion field. Then, we introduce an auto-selecting approach to detect the moving objects for dynamic-aware learning automatically. Empirical experiments demonstrate that our method can achieve the state-of-the-art performance on the KITTI benchmark.
\end{abstract}



\section{Introduction}
\label{sec:intro}

	Inferring the 3D structure and camera's ego-motion from videos is significant in 3D computer vision tasks, especially for robotics, autonomous driving, and augmented reality. The traditional methods take the idea of structure-from-motion, using the stereo matching to recover the depth map, and solving the essential matrix by the perspective-n-point (PnP) \citep{10.1145/358669.358692} method. These geometric methods usually rely on the hand-crafted features with correspondence search and epipolar geometry. Recently, learning-based depth estimation \citep{eigen_NIPS2014,xu2017multi,xu2018structured,FuCVPR18-DORN} and ego-motion estimation \citep{conf/iccv/KendallGC15,Wang2017DeepVOTE,8753500,curriculum-vo-2019} methods have emerged and achieved great success. However, the requirement of the high-precision ground-truth limits their flexibility and transportability. To utilize the pervasive unlabeled monocular videos, Zhou et al. \citep{zhou_unsupervised_2017} propose to learn depth and ego-motion jointly from videos in a self-supervised manner. Along this line, many works have put forward many improvements in consistency constraints~ \citep{mahjourian2018unsupervised, bian2019depth} and awareness of occlusion \citep{godard_digging_2018}. Another remaining challenge is moving objects, which can destroy the scene's rigid global transformation caused by the camera's ego-motion.

%


To tackle the issue of the moving objects, most previous works \citep{without_sensors_2019,Gordon_2019_ICCV} used a pre-trained segmentation model to preprocess the data, and then separately estimated the movements of the individuals. The pre-trained model, like Mask R-CNN \citep{he2017maskrcnn}, is always bulky and introduces the exterior annotated knowledge, making the possible online adaptation \cite{Li_2020_CVPR} difficult, which is one of the main advantages of self-supervised learning. In this paper, we aim to design a dynamic-aware learning system under the pure self-supervision from monocular videos. Since Gordon et al. \citep{Gordon_2019_ICCV} have demonstrated that the CNN-based model can estimate the camera's ego-motion and the motion field of selected regions, monocular videos naturally contain all features required to complete these two estimations. We note that ego-motion estimation, or called visual odometry, relies on static elements in the scene, while dynamic features contribute to motion field estimation for moving objects.

In order to distinguish and exploit the static and dynamic features of the scene, we propose an attentional separation-and-aggregation network (ASANet). There are two weights-sharing feature propagation paths inside the ASANet, corresponding to the static and dynamic characteristics. In each layer of each path, the features are first decomposed into static and dynamic elements using soft attention, and then the decomposed static and dynamic features from two paths are aggregated into a new pair of features and fed into the subsequent layers. Through multiple separation-and-aggregation operations, ASANet finally outputs pure static and dynamic features. Leveraging the ASANet as the encoder, our extended MotionNet can simultaneously estimate the ego-motion of the camera and the scene's motion field by two separate decoders. Also, we design an auto-selecting mechanism to select the appropriate transformation to build supervision automatically, i.e., using ego-motion for projecting stationary elements and motion field for transforming the moving objects. We conduct extensive experiments to verify our method's performance and compare our method with the existing methods, showing that our method can achieve better performance at a lower cost.


Our contributions in this work can be summarized as follows:
\vspace*{-1mm}
\begin{itemize}[leftmargin=*]
    \item We propose an Attentional Separation-and-Aggregation Network (ASANet), which can extract the static and dynamic features simultaneously from stacked images.

    \item We propose an auto-selecting approach to eliminate the moving objects issue in a self-supervised manner. To our best knowledge, we are the first to alleviate effect from arbitrary moving objects in self-supervised depth-pose learning from monocular videos without any exterior knowledge.

    \item Our method achieves state-of-the-art performance on the depth estimation and end-to-end visual odometry tasks of the KITTI benchmark \citep{Geiger2013IJRR,Geiger2012CVPR} with only a small additional cost.
\end{itemize}

\section{Related Work}
\label{sec:related_work}

	\subsection{Supervised learning for depth and ego-motion}

Supervised methods use precise ground-truth to train neural networks, and these ground-truth data often require expensive instruments for measurement, such as LiDAR used to acquire depth and Inertial Navigation Systems (INS) to record poses. Eigen et al. \cite{eigen_NIPS2014} proposed the first CNN-based model to estimate the dense depth from a single raw image. After that, some subsequent works tried some methods to help the model learn better, such as using conditional random fields (CRFs) to fuse multi-scale information \citep{xu2017multi}, introducing structural attention \citep{xu2018structured}, or treating depth estimation as an ordered regression task \citep{FuCVPR18-DORN}. The ego-motion estimation is more famous as visual odometry (VO), which measures the agent's posture transformation from two views. Since VO is a typical sequence-to-sequence task, the recurrent units are widely used in the existing supervised VO methods. Wang et al. \citep{Wang2017DeepVOTE} proposed the first end-to-end learning-based VO approach by leveraging recurrent convolutional neural networks (RCNNs). Jiao et al. \citep{Jiao_2018_ECCV} and Xue et al. \citep{DBLP:conf/cvpr/XueWLWWZ19} improved the model architecture to make better use of temporal geometric relationships. Despite the rapid evolution of these supervised methods, the demand for high-precision ground-truth limits their further development.

\subsection{Self-supervised joint depth-pose learning}

Self-supervised joint depth-pose learning is an emerging topic in recent years, which links depth and ego-motion via structure-from-motion (SfM). Zhou et al. \citep{zhou_unsupervised_2017} proposed a pioneering work in this area. On this basis, some of the follow-up works put forward new consistency constraints~ \citep{mahjourian2018unsupervised,bian2019depth}, and some added occlusion processing in the training framework \citep{godard_digging_2018,zhao2020towards}. Moving objects are also a concern in this genre because they are not satisfied with the transformation calculated by ego-motion. The auto-masking approach proposed by Godard et al. \citep{godard_digging_2018} can avoid the influence of objects that are relatively stationary with the agent. Casser et al. \citep{without_sensors_2019} and Gordon et al. \citep{Gordon_2019_ICCV} proposed to segment all instances in advance by a pre-trained Mask R-CNN \citep{he2017maskrcnn} model and then to estimate the motion field for these individuals. As a similar self-supervised video learning task, self-supervised scene flow learning is also facing moving objects, where some works \citep{lee2019learning,Hur:2020:SSM} use residual flow to narrow the gap caused by moving objects. Unlike these methods, our method uses the attention mechanism to explore the dynamic features of the scene directly and automatically adapts to the static and dynamic elements during training.

On the other hand, some recent works attempted to incorporate geometric VO into the self-supervised manner. Zhen et al. \citep{zhan_visual_2020} first proposed to utilize a FlowNet to predict bi-directional optical flows and samples corresponding points from flows with least forward-backward flow inconsistency to solve fundamental matrix. Zhao et al. \cite{zhao2020towards} further introduced the self-supervised manner to this paradigm and enhanced the generalization ability. However, these geometric VO methods all spend most of their time on feature extraction (or flow estimation) and matching, which makes the method difficult to achieve high running speed. Despite a performance gap from geometric VO, end-to-end VO (or called direct VO) has good potential and can run in real-time.

	\begin{figure}
		\centering
		\includegraphics[width=\textwidth]{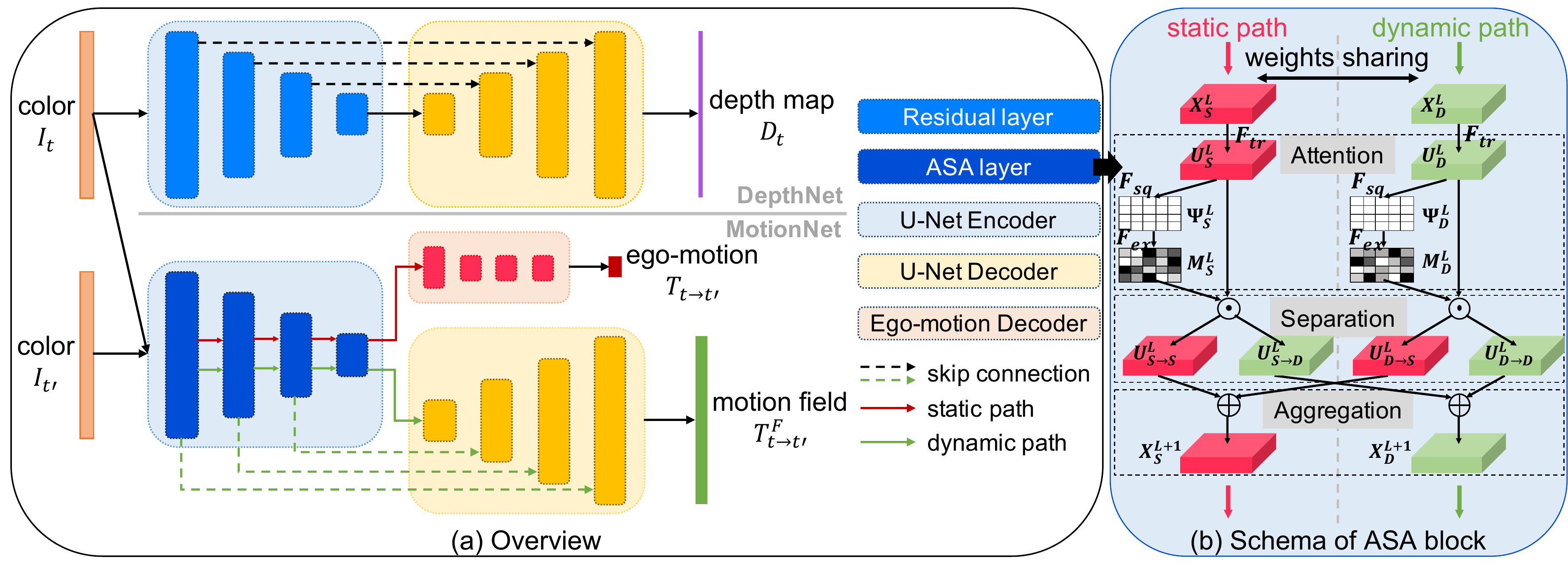}
		\vspace*{-5mm}
		\caption{\label{fig:framework}\textbf{(a) Overview of the joint learning framework:} a DepthNet in the shape of U-Net and a MotionNet composed of ASANet encoder and two separate decoders. \textbf{(b) Schema of the proposed ASA block}, where the red and green volumes represent static and dynamic features, respectively.}
	\end{figure}


\section{Methodology}
\label{sec:methods}
	
	In this section, we describe our proposed Attentional Separation-and-Aggregation Network (ASANet) and the details of the training process. The overall framework of our joint learning system is depicted in \cref{fig:framework}. Following Godard et al. \citep{godard_digging_2018}, we take the ResNet \citep{DBLP:conf/cvpr/HeZRS16} to encoder the input image to leverage the pretraining on ImageNet \citep{imagenet_cvpr09}, and then use the U-shaped decoder to reproject the coded feature into the inverse depth, or called disparity. Besides, we extend the commonly used PoseNet to MotionNet, which estimates the ego-motion from static features and estimates the motion field from dynamics.
	
	\subsection{Attentional Separation-and-Aggregation Network}
	\label{sec:aan}

As aforementioned, only part of features between two frames contributes to the ego-motion estimation, while the others represent the dynamic characteristics good for handling moving objects. We propose an Attentional Separation-and-Aggregation Network (ASANet) to separate and extract static and dynamic features from adjacent frames. In detail, we design a novel ASA block that uses spatial soft attention in the feature space to weight the static elements and therefore can use the opposite attention to weight the dynamic elements in the meantime. 




We diagram the schema of our proposed ASA block in \cref{fig:framework} (b), which contains three main operations: attention generation, feature separation, and aggregation. An ASA block receives two features (static and dynamic) from the last block, notated as $X_S^L$ and $X_D^L$. Each of the two features generates a propagation path and undergoes the following operations. For any given transformation $F_{tr}$, one of the input features $X_i^{L}$  converts to the intermediate feature $U_i^L$, where $U_i^L\in \mathbb{R}^{H\times W\times C}$, and $i\in \{S, D\}$ indicates which feature path is current. In our implementation, we take the non-identity branch of a residual block as $F_{tr}$ like in SE-ResNet \citep{DBLP:conf/cvpr/HuSS18}. Then for \textbf{(a) attention generation}, we first squeeze the feature volume $U_i^L$ at the channel dimension, denoted as $F_{sq}$ in \cref{fig:framework}, to focus on the spatial relationships, thus we can get a spatial correspondence descriptor $\Psi_i^L\in\mathbb{R}^{H\times W\times 1}$. Thereafter, a set of convolutions $F_{ex}$ process the descriptor $\Psi_i^L$ to produce a probability map $M_i^L$, each element of which expresses how likely it describes the static characteristics. For \textbf{(b) feature separation}, we apply the static-likely mask $M_i^L$ and the dynamic-likely mask $(1-M_i^L)$ to $U_i^L$, obtaining two new separated features $U_{i\to S}^L$ and $U_{i\to D}^L$. At the end of an ASA block, we \textbf{(c) aggregate} two new pairs of features from two paths, $X_S^{L+1}=U_{S\to S}^L+U_{D\to S}^L$ and $X_D^{L+1}=U_{S\to D}^L+U_{D\to D}^L$, and finally output the purer static and dynamic features to the rest of the network.

Using an ASANet as the encoder, our MotionNet feeds the extracted static and dynamic features into two independent decoders to regress the ego-motion and dynamic motion field, respectively, among which the motion field decoder and the ASANet constitute a U-shaped structure. Through the skip connections, the impure dynamic features from the shallow layers provide the necessary structural information and some residual static features for restoring the scene's motion field.


	\subsection{Self-supervised objective}
	\label{sec:backgound}

In line with SfMLearner \citep{zhou_unsupervised_2017}, we formulate the self-supervised depth-pose learning as novel view synthesis. Let's denote $<I_{t-1},I_t,I_{t+1}>$ as a training snippet with $I_t$ being the target view and the rest being the source view $I_{t'}(t'\neq t)$. The DepthNet estimates the disparity $D_t$ from a single image $I_t$ and then restores it to the dense depth map by $\hat{D}_t=1/D_t$, while the MotionNet estimates the ego-motion $T_{t\to t'}$ and the motion field $T^F_{t\to t'}$ from the stacked images. Then the estimated depth map $D$ and relative pose $\mathcal{T}$ can warp one view into another according to the epipolar geometry:
\begin{equation}
\label{eq:view-synthesis}
    p_{t'} \sim K\mathcal{T}\hat{D}_t(p_t)K^{-1}p_t
\end{equation}
where $K$ denotes the camera intrinsics, and $p_t, p_{t'}$ are the homogeneous coordinates of pixels in $I_t$ and $I_{t'}$. $\mathcal{T}$ is a short version of $\mathcal{T}_{t\to t'}(p_t)$ for the sake of simplicity, equal to either the ego-motion $T_{t\to t'}$ or one element of the motion field $T^F_{t\to t'}(p_t)$, simply notated as $T$ and $T^F$ below. Utilizing the view synthesis, we can optimize our models by minimizing the following consistency errors.

\textbf{Photometric consistency loss} measures the difference between the target image $I_t$ and the synthesized view $\hat{I}_t^{t'}=\mathcal{F}(I_{t'},\hat{D}_t,\mathcal{T})$, being a combination of a L1 and SSIM \citep{1284395} loss, where $\mathcal{F}$ is the warping operation. Then the photometric error ($pe$) is defined as
\begin{equation}
    \label{eq:photometric}
    pe(I_t,\hat{I}_t^{t'}) = \frac{\alpha}{2}(1-SSIM(I_t,\hat{I}_t^{t'})) + (1-\alpha)||I_t-\hat{I}_t^{t'}||_1
\end{equation}
Inspired by Godard et al. \citep{godard_digging_2018}, we compute the per-pixel minimum error across all source views instead of averaging them and incorporate the auto-masking term (\cref{eq:auto-mask}), to overcome the issues of out-of-view pixels and occlusions.
\begin{equation}
    \label{eq:auto-mask}
    \mathcal{M}_{auto} = \llbracket\min_{t'}{pe(I_t, \hat{I}_t^{t'})} < \min_{t'}{pe(I_t,I_{t'})}\rrbracket
\end{equation}
\begin{equation}
    \label{eq:ppm-photometric}
    \mathcal{L}_{pe}=\mathcal{M}_{auto}\odot\min_{t'}{pe(I_t,\hat{I}_t^{t'})}
\end{equation}
where $\llbracket\cdot\rrbracket$ indicates the Iverson bracket and $\odot$ is the element-wise product.

\textbf{Geometric consistency loss} is proposed in  \citep{bian_unsupervised_2019} to ensure a consistent scale between different training snippets. Similar to photometric consistency, the warping operation is applied but on the depth map rather than the RGB image. We can compute the geometric error ($ge$) by
\begin{equation}
    \label{eq:geometric}
    ge(\hat{D}_t,\hat{D}_{t'},\mathcal{T}) = \frac{|\hat{D}_{t'}^t-\hat{D}_{t'}'|}{\hat{D}_{t'}^t+\hat{D}_{t'}'}
\end{equation}
where $\hat{D}_{t'}^t=\mathcal{F}(\hat{D}_t,\mathcal{T})$ is the warped depth map from $\hat{D}_{t}$ and $\hat{D}_{t'}'$ is the interpolated depth map of $I_{t'}$ aligned with the warping flow. Here, we also compute the per-pixel minimum error instead of the average.
\begin{equation}
    \label{eq:ppm-geometric}
    \mathcal{L}_{ge}=\mathcal{M}_{auto}\odot\min_{t'}{ge(\hat{D}_t,\hat{D}_{t'},\mathcal{T})}
\end{equation}
\textbf{Smoothness regularization} is commonly used in previous works \citep{zhou_unsupervised_2017,godard_digging_2018,lee2019learning}. We adopt two edge-aware smoothness losses here, where one is a first-order term to regularize the depth maps \citep{zhou_unsupervised_2017,godard_digging_2018}, and the other is a second-order term for the motion field \citep{lee2019learning}.
\begin{equation}
    \label{eq:ds}
    \mathcal{L}_{ds}=|\nabla_x^1D_t^*|e^{-|\nabla_x^1I_t|}+|\nabla_y^1D_t^*|e^{-|\nabla_y^1I_t|}
\end{equation}
\begin{equation}
    \label{eq:fs}
    \mathcal{L}_{fs}=|\nabla_x^2F_{t\to t'}|e^{-|\nabla_x^2\hat{D}_t|}+|\nabla_y^2F_{t\to t'}|e^{-|\nabla_y^2\hat{D}_t|}
\end{equation}
where $D_t^*=D_t/\bar{D}_t$ is the mean-normalized disparity map, $\hat{D}_t=1/D_t$ is the restored depth map, and $F_{t\to t'}$ is the warping flow derived from the epipolar projection, an intermediate for smoothing the motion field:
\begin{equation}
    \label{eq:flow}
    F_{t\to t'}(p_t)=KT^F_{t\to t'}(p_t)\hat{D}_t(p_t)K^{-1}p_t-p_t
\end{equation}

	\begin{figure}
		\centering
		\subfigure[Sample1]{\includegraphics[width=0.4\textwidth]{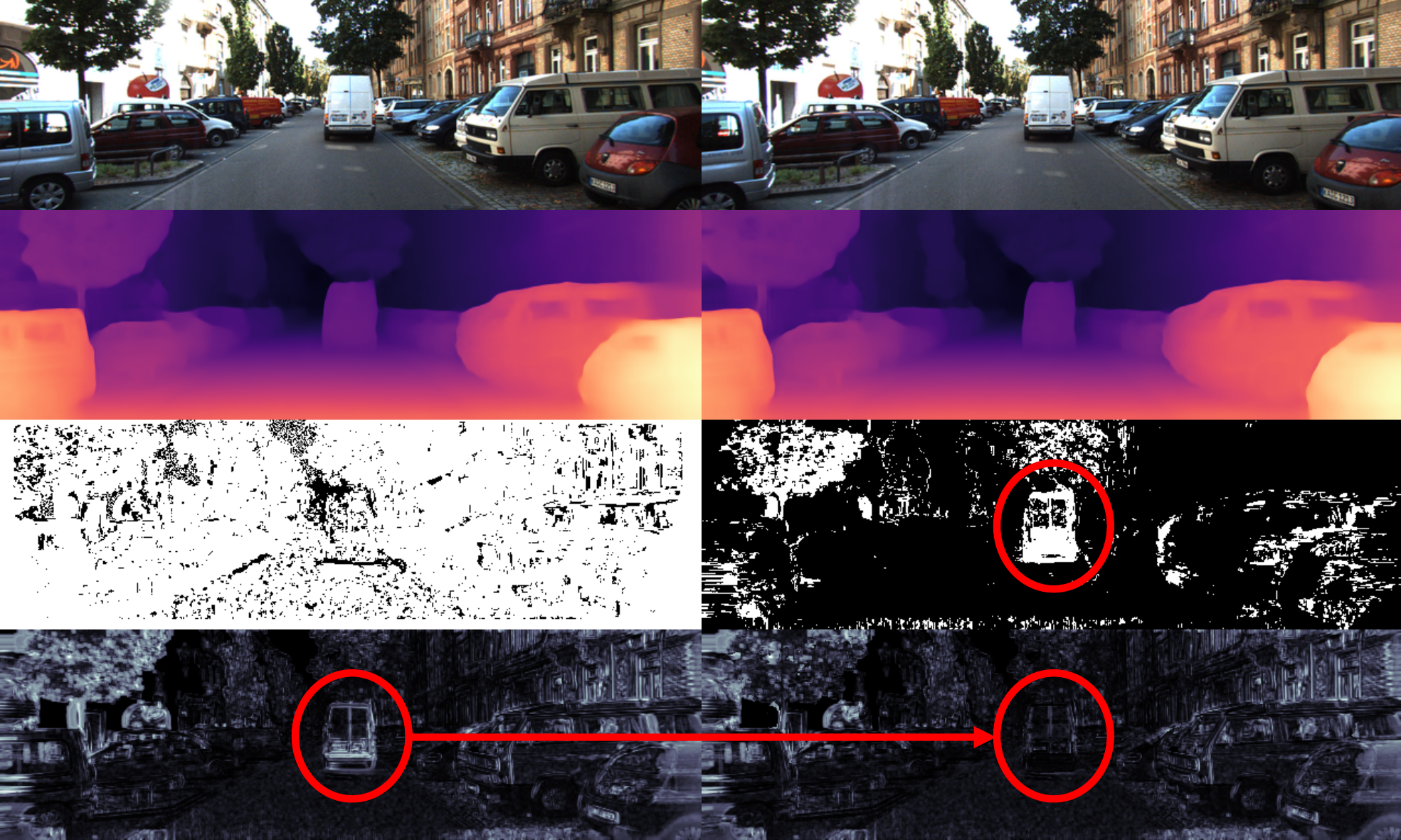}}
		\hspace{0.025\textwidth}
		\subfigure[Sample2]{\includegraphics[width=0.4\textwidth]{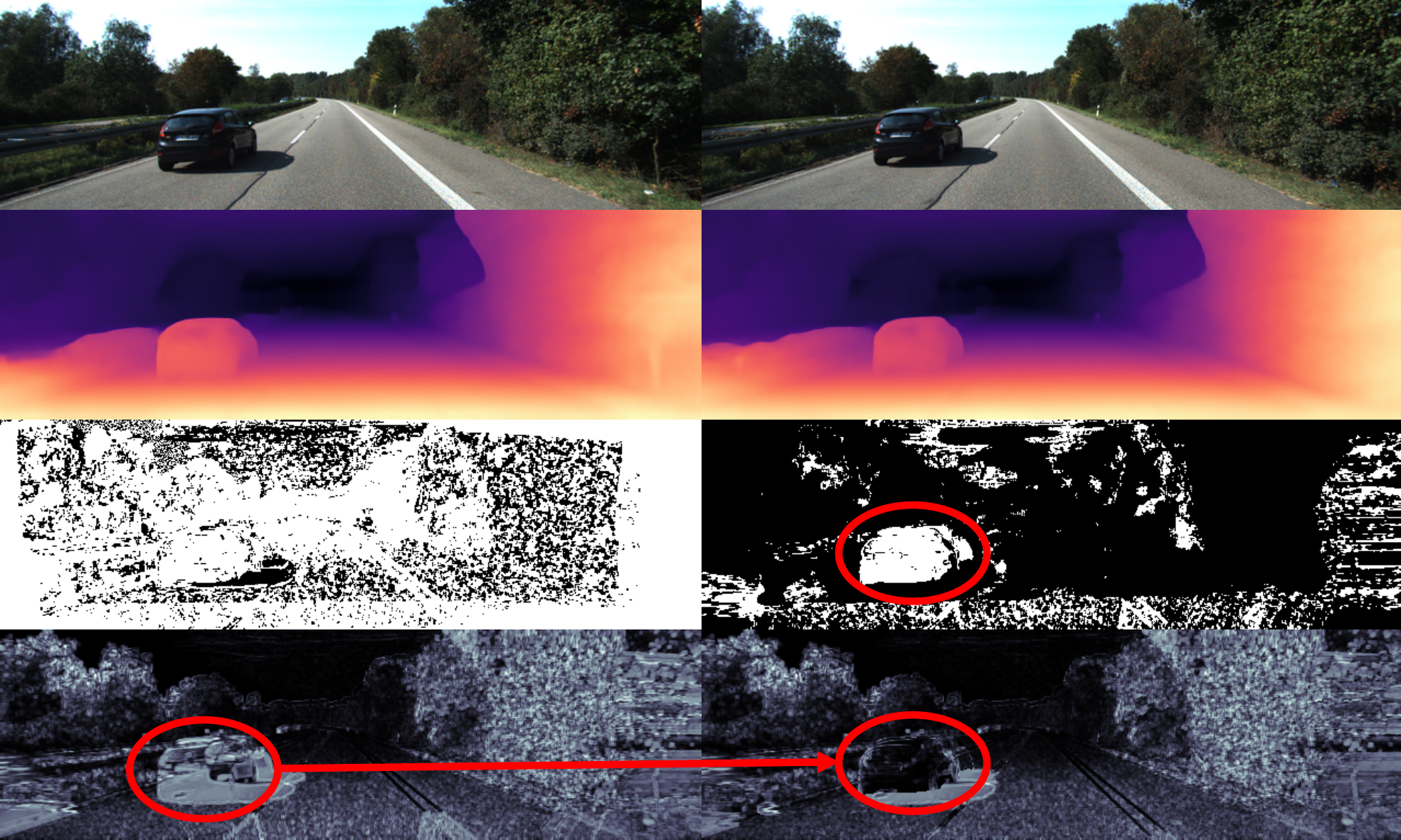}}
		\vspace*{-4mm}
		\caption{\label{fig:visual}\textbf{Visual results to show the effect of our dynamic-aware learning\protect\footnotemark[1].} The first two rows are sample images and corresponding estimated depth maps. The third row shows the results of auto-masking $\mathcal{M}_{auto}$ and our proposed auto-selecting $\mathcal{M}_{dynamic}$ approach. The last row illustrates the photometric error maps before and after the dynamic pixel-wise projection is incorporated into the rigid global projection via our auto-selecting mechanism.}
	\end{figure}

	\subsection{Dynamic-aware learning}
	\label{sec:dynamic}
	Dynamic awareness is critical for view synthesis, because the epipolar projection produced by camera motion is invalid for moving objects. To solve this problem, we propose an auto-selecting mechanism that automatically selects the transformation source for each point by comparing the consistency error from ego-motion and that from the motion field. We can calculate the selective mask according to the following formula
\begin{equation}
    \label{eq:dynamic-mask}
    \mathcal{M}_{dynamic}=\llbracket pe(I_t,\mathcal{F}(*,T)) > \eta\cdot pe(I_t,\mathcal{F}(*,T^F))\rrbracket\odot\llbracket ge(*,T) > \eta\cdot ge(*,T^F)\rrbracket
\end{equation}
where $*$ represents other inputs required, and $\eta$ is a scale factor for robust and stable selection, which we set as 1.2 to perfer the ego-motion. Then we can modify the above consistency losses as
\begin{equation}
    \label{eq:dynamic-error}
    \tilde{\mathcal{E}}(*,T,T^F)=\mathcal{M}_{dynamic}\odot \mathcal{E}(*,T^F) + (1-\mathcal{M}_{dynamic})\odot \mathcal{E}(*,T), \mathcal{E}\in\{pe,ge\}
\end{equation}
\begin{equation}
    \label{eq:dynamic-p}
    \mathcal{L}_{pe}^M=\mathcal{M}_{auto}\odot\min_{t'}\tilde{pe}(*,T,T^F)
\end{equation}
\begin{equation}
    \label{eq:dynamic-g}
    \mathcal{L}_{ge}^M=\mathcal{M}_{auto}\odot\min_{t'}\tilde{ge}(*,T,T^F)
\end{equation}

	\footnotetext[1]{You can find more visual results at https://youtu.be/lMBLlQ5NDUU.}

	\subsection{Multi-Phase training}
	\label{sec:3-stage}
	To effectively utilize the proposed ASANet and dynamic-aware learning, we train our networks in three phases, inspired by \citep{lee2019learning}. For clear expression, we use $\Theta_D$, $\Theta_{ASA}$, $\Theta_E$, $\Theta_F$ to represent the DepthNet, the ASANet-encoder, the ego-motion decoder, and the motion field decoder respectively, all of which make up our total framework $\Xi=\{\Theta_D, \Theta_{ASA}, \Theta_E, \Theta_F\}$. And we use these notations as the input parameters to indicate which parts are optimized by a specific objective.


In the first phase, we set aside the motion field decoder of MotionNet, and learn the depth and the camera's ego-motion that establish the rigid global projection for the warping operation. The overall loss in this phase is a weighted sum of photometric, geometric and disparity smoothness loss.
\begin{equation}
    \label{eq:phase1}
    \mathcal{L}_1=\mathcal{L}_{pe}(\Theta_D,\Theta_{ASA},\Theta_E)+\lambda_G \mathcal{L}_{ge}(\Theta_D,\Theta_{ASA},\Theta_E)+\lambda_D \mathcal{L}_{ds}(\Theta_D)
\end{equation}
After initially training the DepthNet and the static path of MotionNet, we freeze them in the second phase and optimize the motion field decoder. We use the estimated motion field to compute a dynamic transformation for each pixel to handle moving objects. Here, we emphasize again that our proposed ASANet can extract both static and dynamic features, the latter of which is used in this phase.
\begin{equation}
    \label{eq:phase2}
    \mathcal{L}_2=\mathcal{L}_{pe}(\Theta_F)+\lambda_G \mathcal{L}_{ge}(\Theta_F)+\lambda_F \mathcal{L}_{fs}(\Theta_F)
\end{equation}
In the last phase, we jointly training all components in our system. We merge the two optimization paths in prior phases through the auto-selecting mechanism described in \cref{sec:backgound}, rather than simply glue them together. Therefore, the objective is defined as
\begin{equation}
    \label{eq:phase3}
    \mathcal{L}_3=\mathcal{L}_{pe}^{M}(\Xi) + \lambda_G \mathcal{L}_{ge}^{M}(\Xi) + \lambda_F \mathcal{L}_{fs}(\Theta_{ASA},\Theta_F) + \lambda_D \mathcal{L}_{ds}(\Theta_D)
\end{equation}


\begin{table}[t]
    \centering
    \resizebox{0.9\textwidth}{!}{
    \begin{tabular}{lcccccccccc}
        \toprule[2pt]
               &     &           &        &\multicolumn{4}{c}{Error metric $\downarrow$} &\multicolumn{3}{c}{Accuracy metric $\uparrow$}\\
        \cmidrule(lr){5-8} \cmidrule(lr){9-11}
        Method &Sup. &Resolution &Dataset &Abs Rel &Sq Rel &RMSE &RMSE log &$\delta<1.25$ &$\delta<1.25^2$ &$\delta<1.25^3$ \\
        \midrule
        Eigen \citep{eigen_NIPS2014} &D &576 x 172 &K &0.203 &1.548 &6.307 &0.282 &0.702 &0.890 &0.890\\
        Kuzniestsov \citep{Kuznietsov_2017_CVPR} &D+S &621 x 187 &K &0.113 &0.741 &4.621 &0.189 &0.862 &0.960 &0.986\\
        DORN \citep{FuCVPR18-DORN} &D &512 x 385 &K &\textbf{0.072} &\textbf{0.307} &\textbf{2.727} &\textbf{0.120} &\textbf{0.932} &\textbf{0.984} &\textbf{0.995}\\
        \midrule
        SfMLearner \citep{zhou_unsupervised_2017} &M &416 x 128 &CS+K &0.198 &1.836 &6.565 &0.275 &0.718 &0.901 &0.960\\
        DF-Net \citep{zou2018dfnet} &M &576 x 160 &CS+K &0.146 &1.182 &5.215 &0.213 &0.818 &0.943 &0.978\\
        Struct2Depth \citep{without_sensors_2019} &M &416 x 128 &K &0.141 &1.026 &5.291 &0.215 &0.816 &0.945 &0.979\\
        CC \citep{DBLP:conf/cvpr/RanjanJBKSWB19} &M &832 x 256 &CS+K &0.139 &1.032 &5.199 &0.213 &0.827 &0.943 &0.977\\
        SC-SfMLearner \citep{bian_unsupervised_2019} &M &832 x 256 &CS+K &0.128 &1.047 &5.234 &0.208 &0.846 &0.947 &0.976\\
        Self-Mono-SF \citep{Hur:2020:SSM} &M &832 x 256 &K &0.125 &0.978 &4.877 &0.208 &0.851 &0.950 &0.978\\
        Monodepth2 (R18) \citep{godard_digging_2018} &M &640 x 192 &K &0.115 &0.903 &4.863 &0.193 &0.877 &0.959 &0.981\\
        TrianFlow \citep{zhao2020towards} &M &832 x 256 &K &0.113 &\textbf{0.704} &\textbf{4.581} &\textbf{0.184} &0.871 &0.961 &\textbf{0.984}\\
        Monodepth2 (R50) \citep{godard_digging_2018} &M &640 x 192 &K &0.111  &0.825  &4.644  &0.187  &0.883  &0.962  &0.982  \\
        PackNet-SfM \citep{packnet} &M &640 x 192 &K &0.111 &0.785 & 4.601 &0.189 &0.878 &0.960 &0.982\\
        \midrule
        \textbf{Ours (R18)} &M &640 x 192 &K &   0.112  &   0.866  &   4.693  &   0.189  &   0.881  &   0.961  &   0.981  \\
        \textbf{Ours (R50)} &M &640 x 192 &K &   \textbf{0.108}  &   0.820  &   {4.595}  &   {0.186}  &   \textbf{0.886}  &   \textbf{0.963}  &   {0.982}  \\
        \bottomrule[2pt]
	\end{tabular}}
	\vspace*{2mm}
	\caption{\label{tab:depth-results}\textbf{Quantitative results on the Eigen split of KITTI dataset} for distances up to 80m. For training dataset, K means that the method is trained on the KITTI dataset \citep{Geiger2013IJRR}, while CS+K refers to the models with pre-training on Cityscapes \citep{Cordts2016Cityscapes}. D denotes the depth supervision, and S denotes stereo input pairs. M refers to the methods in the self-supervised learning manner with monocular video clips. (R18) and (R50) refer to the models with ResNet18 or ResNet50 as the backbone. Best results in supervised and self-supervised methods are in bold for highlight.}
	\vspace*{-5mm}
\end{table}

\section{Experiments}
\label{sec:result}

	To validate the contributions of our method, we conduct extensive experiments and describe the results in this section. We train and test our models on the KITTI dataset \citep{Geiger2013IJRR} for depth estimation and on the KITTI odometry dataset \citep{Geiger2012CVPR} for visual odometry, for the fair comparison with existing self-supervised methods. 

	\subsection{Datasets}
	\label{sec:datasets}
	\paragraph{KITTI dataset} The KITTI dataset \citep{Geiger2013IJRR} is a standard benchmark for evaluating depth prediction, with LiDAR readings as the ground-truths for evaluation. We use the data split of Eigen et al. \citep{eigen_NIPS2014} and adopt the pre-processing of Zhou et al. \citep{zhou_unsupervised_2017} to remove stationary frames. This results in 39810 snippets for training, 4424 for validation, and 697 for testing.

\paragraph{KITTI Odometry dataset} KITTI Odometry dataset \citep{Geiger2012CVPR} contains 22 stereo sequences with 11 sequences having accessible ground-truth camera poses. Consistently with the literature~ \citep{zhou_unsupervised_2017,godard_digging_2018,zhao2020towards,zhan_visual_2020}, we train our models on sequences 00-08 and evaluate on sequences 09 and 10 with the length of 1591 and 1201 respectively. Then we have 36671 training triplets and 4075 for validation.

	\subsection{Implementation details}
	\label{sec:detail}
	In our framework, the DepthNet is a U-shaped architecture as in  \citep{godard_digging_2018}. The MotionNet is as described in \cref{sec:aan}. The two networks both use ResNet \citep{He2015DeepRL} as the backbone which is pretrained on ImageNet~ \cite{imagenet_cvpr09}.

Our models are implemented in PyTorch \citep{NIPS2019_9015}. For all experiments, we train our models with 3-frame sequences $\{I_{t-1},I_t,I_{t+1}\}$ in one step, and resize the input images to $640\times 192$ during both training and testing. We use the Adam optimizer \citep{kingma2014adam} with $\beta_1=0.9$ and $\beta_2=0.999$. The loss weightings are $[\lambda_G,\lambda_F,\lambda_D]=[0.1,0.001,0.001]$. To evaluate the performance of our proposed framework, we train our models with the multi-phase schedule (\cref{sec:3-stage}), where each phase is for 10 epochs. Since the optimized components in the first two phases have no overlap, we train each component in our system for 20 epochs in total with a batch size of 4. We set the initial learning rate as $10^{-4}$, which is decayed by 10 for refinement after 15 total training epochs.

	\subsection{Evaluation of depth estimation}
	\label{sec:depth-results}
	For depth estimation, we train our models in the three-stage schedule on preprocessed KITTI dataset~ \citep{Geiger2013IJRR}, as described in \cref{sec:3-stage}. Following Godard et al. \citep{godard_digging_2018}, we capture the depth up to 80m and scale the depth to align the per-image median with the median of ground-truth during evaluation. We compare our method with the existing methods in \cref{tab:depth-results}, which shows that our method can achieve state-of-the-art performance. Despite a performance gain for depth estimation, we use the standard U-shaped DepthNet without any additional computation cost, which will be illustrated in \cref{sec:ablation}. That's to say, if we incorporate an improved DepthNet, such as PackNet proposed in \citep{packnet}, we can further boost the performance of our method. In other words, our method can be integrated with other self-supervised methods by replacing the traditional PoseNet for ego-motion estimation by our MotionNet and adopting the dynamic-aware auto-selecting mechanism during training.

To verify the effect of our proposed auto-selecting mechanism, we visualize the qualitative samples in \cref{fig:visual}, consisting of sample images, estimated depth, masks ($\mathcal{M}_{auto}$ and $\mathcal{M}_{dynamic}$) from auto-masking and auto-selecting approaches, and photometric error maps before and after fusing the rigid global projection with the dynamic pixel-wise projection. From the third row in \cref{fig:visual}, we can see that the auto-masking $\mathcal{M}_{auto}$ can reduce the influence of the sky and other areas where the texture change is not obvious, while our dynamic-aware mask $\mathcal{M}_{dynamic}$ can effectively identify the moving objects in the scene. The change of the photometric error maps in the last line can explain that through our auto-selecting mechanism, the redundant errors caused by moving objects can be reduced, thereby avoiding the negative effects of these outliers. Besides, for areas with more detailed textures, such as leaf and bush in Samples 1 and 2, the auto-selecting mechanism can also provide better supervision.


	\subsection{Evaluation of visual odometry}
	\label{sec:vo-results}
	\begin{table}
    \centering
    \resizebox{0.8\textwidth}{!}{
    \begin{tabular}{llccccc}
        \toprule[2pt]
           & &   &\multicolumn{2}{c}{Seq. 09} &\multicolumn{2}{c}{Seq. 10}\\
        \cmidrule(lr){4-5} \cmidrule(lr){6-7}
        Type &Method &Latency &$t_{err} (\%)$ &$r_{err} (\circ/100m)$ &$t_{err} (\%)$ &$r_{err} (\circ/100m)$\\
        \midrule
        \multirow{3}{*}{Geometric VO} &ORB-SLAM2 (w/o LC) \citep{murORB2} &- &9.31 &0.26 &\textbf{2.66} &0.39\\
        &ORB-SLAM2 (w/ LC) \citep{murORB2} &- &\textbf{2.84} &\textbf{0.25} &2.67 &\textbf{0.38}\\
        &TrianFlow \citep{zhao2020towards} $^\ddagger$ &120.25 ms &6.93 &0.44 &4.66 &0.62\\
        \midrule
        \multirow{4}{*}{End-to-end VO} &SfMLearner \citep{zhou_unsupervised_2017} &1.68 ms &11.32 &4.07 &15.25 &4.06\\
        &Depth-VO-Feat \citep{Zhan_2018_CVPR} &2.11 ms &9.07 &3.80 &9.60 &\textbf{3.41}\\
        &SC-SfMLearner \citep{bian_unsupervised_2019} $^\dagger$ &9.38 ms &7.64 &2.19 &10.74 &4.58\\
        &Ours (R18) &20.68 ms &\textbf{5.33} &\textbf{2.13} &\textbf{8.79}  &3.46\\
        \bottomrule[2pt]
    \end{tabular}}
    \vspace*{1mm}
    \caption{\label{tab:vo-results}\textbf{Quantitative results on KITTI Odometry Sequences 09 and 10.} The latency for one inference, average translation ($t_{err}$), and rotation errors ($r_{err}$) are reported. $^\dagger$ indicates that we use their updated results in \citep{zhan_visual_2020}. $^\ddagger$ indicates when testing its running time, we only report the time it uses a PWC-Net~ \citep{Sun2018PWC-Net} to infer the forward and backward optical flow. LC=Loop Closure.}
    \vspace*{-3mm}
\end{table}
\begin{figure}
    \centering
    \subfigure[Seq. 09]{\includegraphics[height=0.3\textwidth]{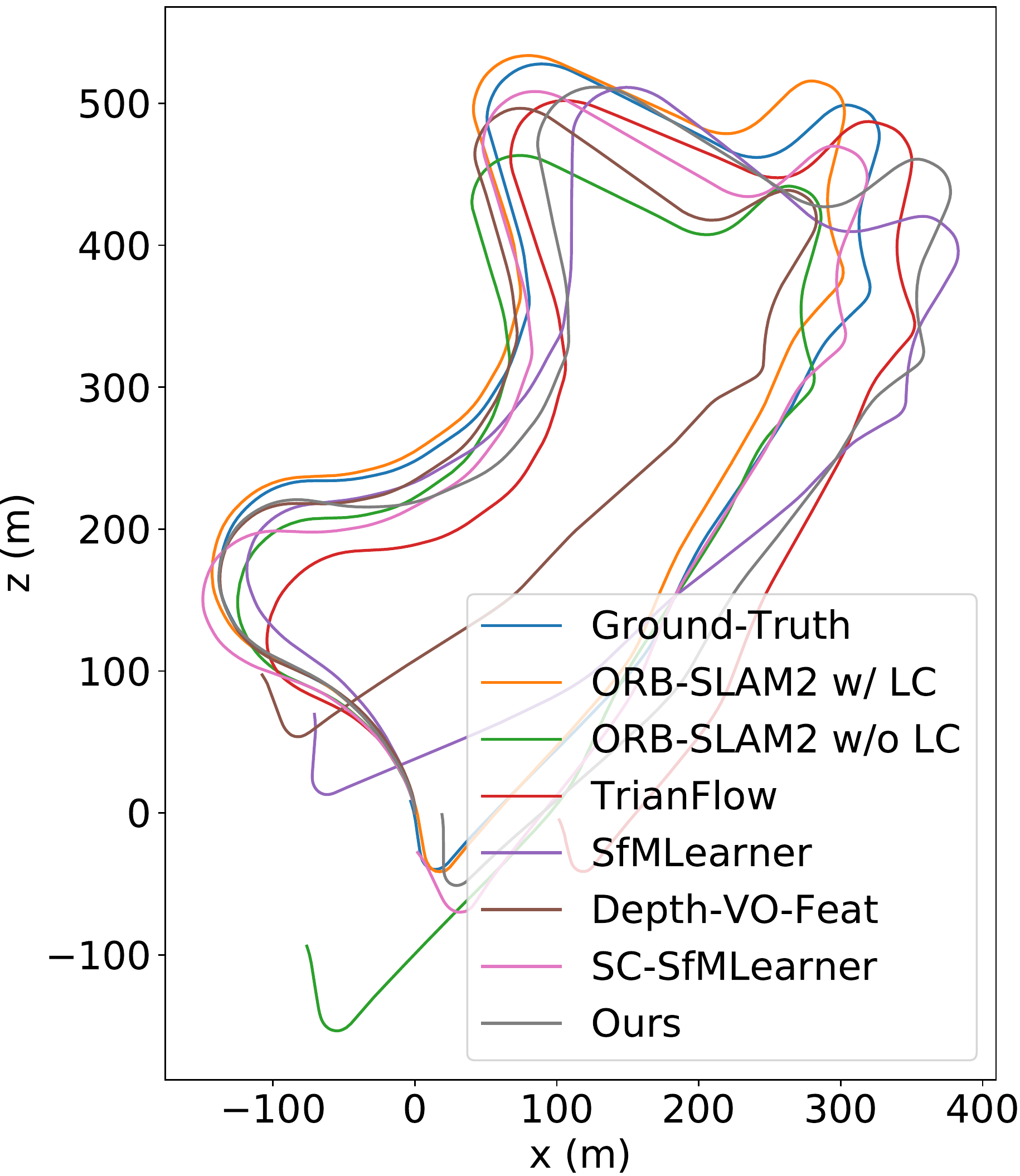}}
    \hspace{0.05\textwidth}
    \subfigure[Seq. 10]{\includegraphics[height=0.3\textwidth]{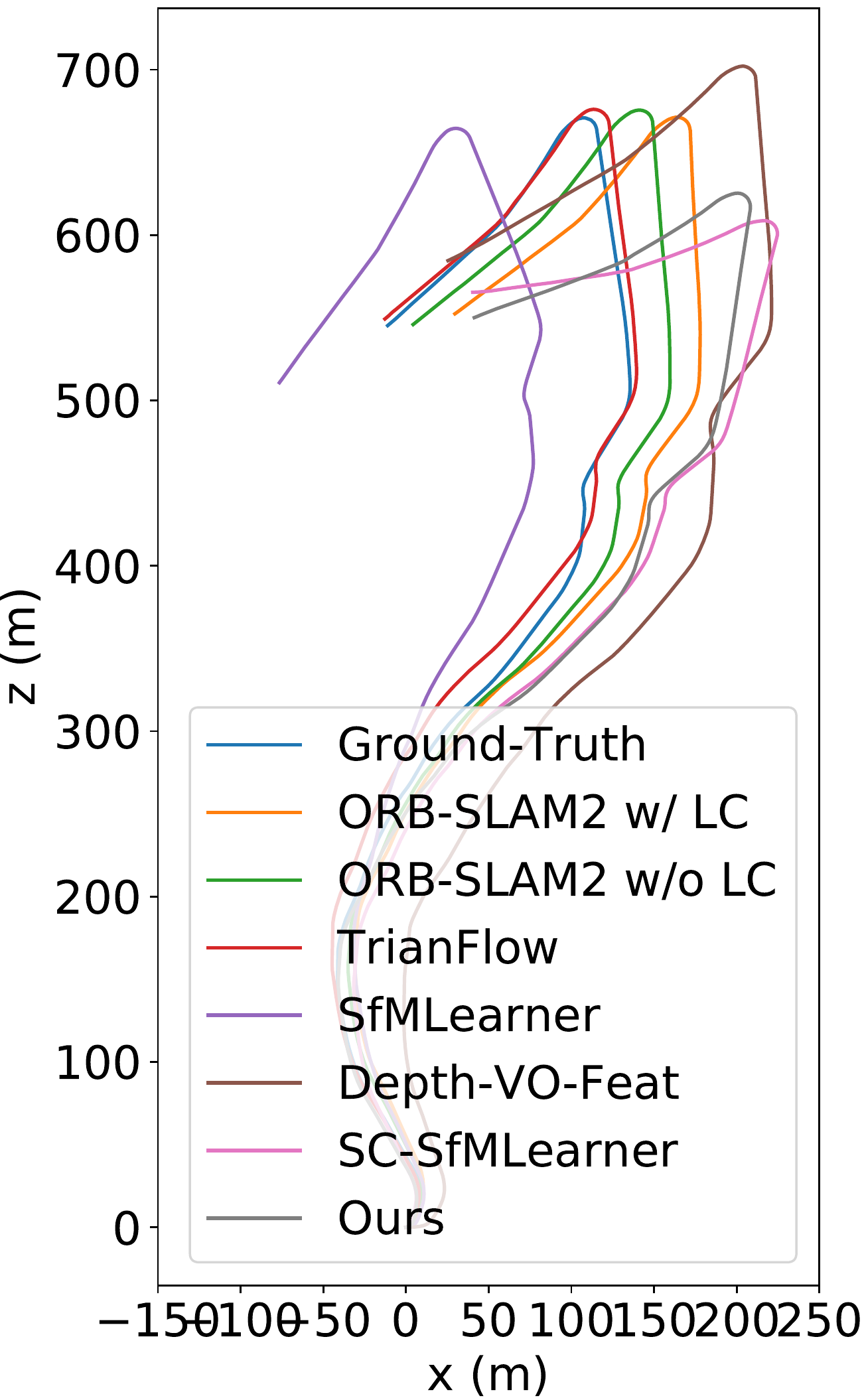}}
    \vspace*{-3mm}
    \caption{\label{fig:vo-fig}\textbf{Qualitative trajectory results on KITTI Odometry Sequences 09 and 10.}}
\end{figure}

We use the same experimental settings as the depth estimation and train our networks on the KITTI odometry sequences 00-08 according to only the first phase in \cref{sec:3-stage}, to compare our methods with some recent self-supervised VO methods on sequences 09 and 10. We scale and use a 7-DoF transformation to align our estimated poses with the ground-truth as Zhan et al. \citep{zhan_visual_2020} did. As a robot localization module that needs to be run in real-time, the model's run-time latency has a great influence on the practical application. So we compare our methods with the existing methods on both performance and the run-time latency in \cref{tab:vo-results}. To test the latency of the model, we forward each model a thousand times on an Nvidia GeForce RTX 2080Ti GPU and take the average running time as the results. It is worth mentioning that in the inference process of VO, we only need to run ASANet-encoder and ego-motion decoder of our MotionNet. Especially, TrianFlow \citep{zhao2020towards} is a two-stage method that first uses PWC-Net \citep{Sun2018PWC-Net} to estimate forward and backward optical flow, then utilizes the geometric VO methods. In our experiments, we only report the running time for its two inferences for forward and backward optical flow estimation. Compared with ORB-SLAM2 \citep{murORB2}, TrianFlow uses FlowNet to replace the process of calculating geometric feature descriptors and matching points, which is still a geometry-based VO, so we group them together. It can be seen from the comparison in \cref{tab:vo-results} that our method can achieve the best in one-stage end-to-end VO methods, and maintain a real-time operating speed of 50 fps. The visualized trajectories on KITTI Odometry sequences 09 and 10 are shown in \cref{fig:vo-fig}.

	\subsection{Ablation study}
	\label{sec:ablation}
	\begin{table}
    \centering
    \resizebox{0.8\textwidth}{!}{
    \begin{tabular}{lccccc}
        \toprule[2pt]
        Method &\#Params &Abs Rel &Sq Rel &$\delta_{1.25}$\\
        \midrule
        PackNet-SfM (D=4) \citep{packnet} &78.49 M &0.113 &0.818 &0.875\\
        PackNet-SfM (D=8) \citep{packnet} &128.29 M &0.111 &\textbf{0.785} &0.878\\
        \midrule
        Monodepth2 (R18) \citep{godard_digging_2018} &14.84 M &0.115 &0.903 &0.877\\
        + $\mathcal{L}_{ge}$ (\cref{eq:ppm-geometric}) &14.84 M &0.114 &0.874 &0.877\\
        + $\mathcal{L}_{ge}$ + ASANet (\cref{sec:aan}) (ours) &14.84 M &0.113 &0.870 &0.878\\
        + $\mathcal{L}_{ge}$ + ASANet + Auto-selecting (\cref{sec:dynamic,sec:3-stage}) (ours) &14.84 M &0.112 &0.867 &0.882\\
        \midrule
        Monodepth2 (R50) \citep{godard_digging_2018} &34.57 M &0.111  &0.825  &0.883\\
        + $\mathcal{L}_{ge}$ (\cref{eq:ppm-geometric}) &34.57 M &0.109  &0.823  &0.885\\
        + $\mathcal{L}_{ge}$ + ASANet (\cref{sec:aan}) (ours) &34.57 M &0.109 &0.824 &0.886\\
        + $\mathcal{L}_{ge}$ + ASANet + Auto-selecting (\cref{sec:dynamic,sec:3-stage}) (ours) &34.57 M &\textbf{0.108} &{0.820} &\textbf{0.886}\\
        \bottomrule[2pt]
    \end{tabular}}
    \vspace*{2.5mm}
    \caption{\label{tab:ablation}\textbf{Ablation study} on standard KITTI benchmark \citep{Geiger2013IJRR} which compares the proposed methods with our baseline, Monodepth2 \citep{godard_digging_2018}, and a state-of-the-art method, PackNet-SfM \citep{packnet}. (R18) and (R50) indicates the backbone of the DepthNet.}
    \vspace*{-7mm}
\end{table}

The performance of our novel self-supervised joint learning system has been shown above. In this section we want to verify the effects of the several improvements we have proposed, including the ASANet (\cref{sec:aan}) and the dynamic-aware learning (\cref{sec:dynamic,sec:3-stage}). We take the Monodepth2~ \citep{godard_digging_2018} as a strong baseline, and the state-of-the-art PackNet-SfM \citep{packnet} as a strong reference. In the experimental setting without the auto-selecting mechanism, we train our models according to the loss function shown in \cref{eq:phase1}, and extend its training time to 20 epochs. For the complete dynamic-aware method that incorporates both ASANet and the auto-selecting mechanism, we use the multi-stage schedule proposed in \cref{sec:3-stage} for training, and each stage lasts for 10 epochs. 


The experimental results are shown in \cref{tab:ablation}, which demonstrates that the system's performance can be improved by adding our proposed ASANet and multi-phase dynamic-aware learning. Compared with PackNet-SfM \citep{packnet}, our method can boost the system's performance while the model parameters are less than half of it. This makes our model more advantageous when deployed on the robot. Besides, we notice that when we use ResNet50 as the DepthNet's backbone, the gain effect of ASANet without auto-selecting is reduced in the depth estimation task, but the auto-selecting mechanism for dynamic objects can still improve the performance of the model.

\section{Conclusion}
\label{sec:conclusion}

In this paper, we mainly track moving objects issue, which is a stumbling block in self-supervised video learning. Unlike many previous methods \citep{without_sensors_2019, Gordon_2019_ICCV,lee2019learning} that use additional information such as segmentation and stereo cues, we propose an Attentional Separation-and-Aggregation Network (ASANet) that can automatically distinguish dynamic and static features in the scene, and design an auto-selecting mechanism to achieve dynamic-aware learning. We conduct extensive experiments to verify the performance of our method. The results show that the auto-selecting mechanism we propose can automatically detect moving objects and select appropriate transformations for reconstruction, thus avoiding the interference of these exceptions to the joint depth-pose learning. Moreover, our method can achieve competitive performance on depth estimation and visual odometry while maintaining a low computational cost compared with the existing state-of-the-art methods. In future work, we can try further to optimize the system from the perspective of depth estimation, such as incorporating the 3D-Packing proposed in PackNet-SfM \citep{packnet}, or by utilizing the RNN structure often used in supervised visual odometry methods \citep{Jiao_2018_ECCV,DBLP:conf/cvpr/XueWLWWZ19} to leverage the historical information in the video.



\clearpage
\acknowledgments{This work is supported by National Key R\&D Program of China (2018YFB0105000). This work is also supported by Meituan-Dianping Group, National Natural Science Foundation of China (No. U19B2019, 61832007, 61621091), Tsinghua EE Xilinx AI Research Fund, Beijing National Research Center for Information Science and Technology (BNRist) and Beijing Innovation Center for Future Chips.}


\bibliography{example}  


\end{document}